# Is Trust Correlated with Explainability In AI? A Meta-Analysis

Zahra Atf, *Member, IEEE,* Peter R. Lewis, *Member, IEEE*

*Abstract*— This study critically examines the commonly held assumption that explicability in artificial intelligence (AI) systems inherently boosts user trust. Utilizing a meta-analytical approach, we conducted a comprehensive examination of the existing literature to explore the relationship between AI explainability and trust. Our analysis, incorporating data from 90 studies, reveals a statistically significant but moderate positive correlation between the explainability of AI systems and the trust they engender among users. This indicates that while explainability contributes to building trust, it is not the sole or predominant factor in this equation. In addition to academic contributions to the field of Explainable AI (XAI), this research highlights its broader socio-technical implications, particularly in promoting accountability and fostering user trust in critical domains such as healthcare and justice. By addressing challenges like algorithmic bias and ethical transparency, the study underscores the need for equitable and sustainable AI adoption. Rather than focusing solely on immediate trust, we emphasize the normative importance of fostering authentic and enduring trustworthiness in AI systems.

## I. INTRODUCTION

HUMANS are increasingly interacting with autonomous and semi-autonomous systems. This trend presents two primary challenges: Machine Ethics and Machine Explainability [1],[2]. While Machine Ethics focuses on setting behavioral guidelines to ensure systems act in morally acceptable ways [3], Machine Explainability ensures that these systems can clarify their actions and explain their choices in a manner that's comprehensible and trustworthy for human users [4].

Explainable Artificial Intelligence (XAI) is a rapidly expanding field driven by concerns surrounding the adoption of high-performance "black-box" AI models and the ethical implications of AI usage. To address the public's apprehensions about AI, research has focused on explainability as a means to enhance trust and acceptance [5].

The provision of explanations can instill emotional confidence in users [6]. However, empirical studies on human-AI interaction have yielded varied outcomes [7], [8], [9], [10] in regard of trust enhancement through explainability.

**Z. Atf** is a visiting researcher with the Ontario Tech University, North Oshawa, Canada (Zahra.Atf@ontariotechu.ca). (Corresponding Autor)

**P. R. Lewis** is with the Faculty of Business and Information, Technology Ontario Tech University, North Oshawa, Canada (peter.lewis@ontariotechu.ca).

The primary objective of this paper is to reexamine the prevalent belief that enhancing explicability in AI systems automatically leads to increased user trust. By meta-analyzing 90 studies from the existing literature, we explore the nuanced relationship between explainability and the degree of trust users place in these systems. Our findings reveal a statistically significant, though moderate, positive correlation—suggesting that while explainability plays a role in building trust, it is not the sole determinant.

While our primary focus remains on non-generative AI contexts, we acknowledge that explainability for Generative AI constitutes a nascent and challenging domain in which theoretical and methodological developments are still emerging. Accordingly, future studies could fruitfully extend this research by exploring the distinctive explainability challenges posed by rapidly evolving generative models.

Our study is grounded in a systematic review framework that employs the PRISMA protocol, ensuring a rigorous, reproducible, and transparent synthesis of evidence while addressing challenges such as algorithmic bias and ethical transparency. The paper is structured as follows. Section III details the methodology, outlining our systematic review process and the application of the PRISMA protocol. Section IV presents the inferential statistics derived from our meta-analysis. Finally, Section V discusses our findings and their broader socio-technical implications.

## II. BACKGROUND

In the healthcare domain, Ghassemi et al [11] contends that current explainability methods are unlikely to fulfill these expectations when it comes to patient-level decision support and it is "a false hope" [11]. The difference between "Explainability" and "Explicability" can be understood both in terms of their scope and their context of use: "Explainability" is an AI-focused term that typically involves providing mechanical or technical explanations of how a model works—often conflated with Interpretability—while "Explicability" is a broader philosophical or epistemological concept [12]. In theoretical and ethical frameworks, we invoke "Explicability" to emphasize accountability by ensuring that users, stakeholders, and those affected by a model's outputs can properly understand and, if needed, challenge its results. In contrast, in practical AI contexts and discussions related to system implementations, we apply "Explainability" to describe the model's inner workings or the reason for its behavior. In other words, "Explicability" extends beyond merely interpreting the model's mechanisms or influencing factors and addresses the responsible and accountable use of AI by providing the necessary context and tools for stakeholders to engage with and scrutinize the system effectively.

Trust is cultivated through unwavering, secure, and foreseeable performance that safeguards user interests without manipulative intent, thereby engendering a dependable relationship between users and systems and facilitating seamless interactions and collaboration [10]. Explainability, transparency, and contestability contribute to building this trust by demonstrating how models function and clarifying the reasons behind their decisions, although sometimes enhancing explainability may come at the cost of reduced privacy [13]. In AI-based decision-support models, particularly in high-stakes domains such as healthcare or autonomous vehicles, trust typically refers to confidence in the model's correct and logical performance under complex data conditions and sensitive outcomes [14]. This technical form of trust represents the relevant part of a broader notion of trust that centers on understanding the model's mechanisms and ensuring accountability in the face of potential errors [15]. Consequently, this study focuses on trust in terms of confidence and predictability, and does not address the emotional or psychological aspects of trust typically observed in human-human relationships, though these can certainly also be relevant in AI systems that present as agents [33].

Unlike conventional technologies that operate under static 'set and forget' rules, AI systems continuously evolve, making it essential to reassess trust over time. Users must remain vigilant, as over-reliance without ongoing evaluation may introduce unforeseen risks or vulnerabilities. Ensuring that trust remains well-founded requires periodic scrutiny and adaptation to emerging challenges in AI decision-making.

XAI aims to make AI systems transparent, meeting societal demands for clear decision-making explanations as their use in complex areas grows [16]. The limited transparency in AI models, combined with issues related to data biases and privacy, present significant challenges for users, developers, and societies at large [7]. Large Language Models (LLMs) can perform well on numerous tasks by utilizing chain-of-thought reasoning (CoT), where they generate step-by-step explanations before delivering a final answer. These CoT explanations might appear to transparently reveal how LLMs solve tasks, potentially increasing safety. However, according to Turpin [8], these explanations may not accurately represent the actual reasons behind the model's predictions, often giving a misleading sense of how decisions are made [8]. However, AI systems, particularly LLMs, can generate incorrect or misleading information, known as "hallucination." This undermines trust as users might rely on inaccurate outputs [9]. Despite efforts to enhance explainability and transparency, the unpredictability of AI necessitates vigilance and improved mechanisms to detect and mitigate hallucinations. Addressing and communicating these risks plays an important role in maintaining trust in AI systems [17].

The study by Agbese et al [10] examines the privacy and trust challenges arising from innovations in artificial intelligence, such as liquid AI, and emphasizes the critical importance of maintaining transparency and accountability in these complex and evolving systems. The research highlights that the decentralized nature of AI algorithms within isomorphic Internet of Things (IoT) frameworks increases the risk of security breaches and unauthorized data manipulation [10].

Artificial intelligence (AI) has seen significant success in real-world applications by learning and extracting patterns from complex data. Machine learning (ML) and deep learning (DL) techniques have been applied to tasks like classification and prediction, especially in critical areas such as finance [18] and healthcare [19]. This success necessitates understanding the often-opaque mechanisms of these models. XAI has emerged to provide methods for clarifying AI systems and their outputs, emphasizing its practical and ethical importance in critical domains [20].

Research in the field of XAI has matured, and the academic community has begun to critically examine the paths and justifications presented in the literature for advocating XAI. Studies have shown that many investigations in this area have failed to incorporate perspectives from disciplines outside of computer science [2]. It has also been emphasized that explanations must be tailored to specific audiences who consume the outputs of models, adding to the complexity of XAI. Furthermore, the importance of interdisciplinary collaborations in developing practical explanations for all stakeholders has been highlighted, revealing that there are few empirical studies on the effectiveness of explanations [1].

The DARPA (Defense Advanced Research Projects Agency) XAI program, officially launched in 2016, aimed to create "a suite of new or modified machine learning techniques that produce explainable models that, when combined with effective explanation techniques, enable end users to understand, appropriately trust, and effectively manage the emerging generation of Artificial Intelligence (AI) systems" [21]. This initiative, which ran from 2017 to 2021, spurred a surge in research on making AI decisions more transparent. According to Adamson, it fundamentally assumed that "researchers [could] demonstrate how AI decisions might be explained," and DARPA emphasized "explainable AI—especially explainable machine learning" as essential for warfighters to collaborate effectively with AI "partners" [22].

Shin [5] conducted a study to explore the impact of AI explainability on user trust and attitudes toward AI. The study considered causability as a precursor to explainability and an important factor in AI algorithms, examining its influence on user perceptions of AI-driven services. The findings revealed the dual roles of causability and explainability in shaping trust and subsequent user behaviors. Providing explanations for recommended news articles instilled user trust, while the causability of their understanding of the explanations boosted users' emotional confidence. Causability played a significant role in shaping what and how to explain, affecting the relevance of explainability features [5]. These results suggest the importance of incorporating causability and explanatory cues in AI systems to enhance trust and enable users to evaluate the quality of explanations.

In a study carried out by Cech & Wagner [23] at Aspern See Stadt in Vienna, they adopted an action-research approach. This involved the local community working together with Aspern. Mobil LAB to co-create a digital platform showing individual CO2 footprints. Interestingly, participants weren't significantly swayed by detailed breakdowns of how their CO2 footprints were calculated. Instead, their trust largely stemmed from the belief that the tool was crafted by researchers, not by a commercial entity aiming for profit. This distinction heightened their trust. While participants recognized and appreciated efforts towards transparency in algorithmic design, they didn't heavily weigh the tool's accountability. Despite valuing the comprehensive explanations of data sources, calculation techniques, and potential limitations, these specifics didn't notably change their overall experience or intent to use this eco-feedback mechanism [24].

Ayoub et al [24] conducted a study that demonstrated the positive impact of adding explanations on participants' trust and willingness to share information. The participants' trust in the model predictions was influenced by the source of information and evidence provided. However, the effect of adding the source and evidence on their trust was uncertain. Participants tended to trust the model predictions more when they had prior knowledge. The model explanation was particularly effective in enhancing trust for false claims compared to true claims [24]. This paradoxical finding suggests that while explanations can enhance trust, they may also lead to over-reliance on incorrect information if not carefully managed. This raises important questions about the nature of trust in AI systems and the ethical implications of relying on explanations that may inadvertently support false claims.

The majority of XAI methods are "post-hoc" and inductive, often struggling to make the complex "black box" of AI systems transparent. As a result, they generally do not describe causal relationships, making it challenging to fully understand and convey causality. Some scholars argue that, due to these limitations, only simpler AI models should be used, as they are easier to explain and understand causally.

In the field of philosophy, normative ethics explores how one should act from a moral standpoint. Within normative ethics, there are four well-known schools of ethical thought: utilitarian, deontological, causability, and virtue ethics. Each of these perspectives offers unique ways of thinking and understanding moral issues. Focusing solely on one of these perspectives could lead to overlooking important contextual nuances that should be clarified, validated, and explained to end users and other stakeholders involved [25]. Utilitarian reasoning focuses on the outcomes of actions, such as the potential benefits of AI recommendations. Deontological thinking prioritizes fundamental principles like rights and responsibilities over just outcomes, valuing human life as a paramount right. Causability emphasizes understanding the cause-and-effect relationships of actions to ensure they lead to predictable and morally justifiable outcomes. Virtue ethics emphasizes adhering to higher moral standards, promoting virtues like honesty and generosity beyond obligation [26].

As XAI explanations aim to offer technical transparency, they should shift their focus towards furnishing practical feedback that facilitates meaningful interactions between end-users and AI systems, allowing for more enriched engagement [27]. There is a divergence between previous research and the prevailing assumption that explanations bolster trust. While certain explanations might indeed enhance trust in particular contexts, this is not a given across all situations. This study seeks to determine if there is a significant relationship between explainability and trust, and to assess whether explainability genuinely boosts trust. To address this, we applied a meta-analysis method and conducted a detailed review of the existing literature.

## III. METHODOLOGY

The process of systematic review can be defined as an examination of the available evidence—based on a specific question—which employs systematic and explicit methods to identify, select, and critically analyze relevant research in order to extract and analyze data from it [28] [29]. Therefore, this methodological process is reproducible, scientific, and transparent, and aims to minimize bias throughout the literature reviews [30]. Systematic reviews are based on transparent protocols that include information such as the questions to be addressed by the study, the population or sample that is the focus of the study, the strategy for identifying relevant studies, and the review criteria used by other studies [28] [31]. In our study, we utilized the PRISMA (Preferred Reporting Items for Systematic Reviews and Meta-Analyses) [32] protocol to ensure a rigorous and comprehensive systematic review. Given the research question, we conducted a search on Scopus, IEEE, and Web of Science limited to a 7 year window between 2017 and 2024 using the following criteria: (TITLE-ABS-KEY ( "Explainability" ) AND TITLE-ABS-KEY ( "trust" ) ) AND ( LIMIT-TO ( SRCTYPE, "j" ) OR LIMIT-TO ( SRCTYPE, "p" ) ) AND ( LIMIT-TO ( SUBJAREA, "COMP" ) ) AND ( LIMIT-TO ( DOCTYPE, "cp" ) OR LIMIT-TO ( DOCTYPE, "ar" ) OR LIMIT-TO ( DOCTYPE, "re" ) OR LIMIT-TO ( DOCTYPE, "cr" ) ).

Building on a substantial body of literature featuring both relevant keywords and consistently rigorous methodologies, these criteria were selected to ensure a reliable foundation for our analysis. Furthermore, to preserve objectivity and mitigate researcher bias in the study selection process, we adhered strictly to the PRISMA protocol [32].

In conformity with this protocol, all studies that did not report the main statistical indicators (such as statistical sample size, validity and reliability, composite reliability, mean and variance, correlation coefficient, etc.) or did not use correct statistical methods, as well as all qualitative and mixed studies that did not report statistical indicators, were excluded. Based on this protocol, out of 1305 (IEEE n =302; Scopus n =510; Web of Science n =493) studies regarding the relationship between explainability and trust, only 90 studies were selected to be analyzed.

Among these studies, 17 focused on data governance and ethical considerations, 22 addressed algorithmic transparency and bias mitigation, 18 explored user interface design for explainability, and the remaining 33 covered a diverse range of topics including human-AI interaction, decision-making processes, and trust dynamics. The AI algorithms employed across these studies predominantly included deep neural networks (25 studies), decision trees (20 studies), support vector machines (15 studies), and ensemble methods (10 studies), highlighting a varied application of machine learning techniques. In terms of explainability methods, 30 studies utilized model-agnostic approaches such as SHAP and LIME, 25 employed intrinsically interpretable models, and 20 integrated visualization-based techniques like Integrated Gradients and Layer-wise Relevance Propagation. The presentation of explainability results within user interfaces was achieved through interactive dashboards (40 studies), narrative-based explanations (25 studies), visualizations including interactive charts and comparative tools (20 studies), and multi-level explanation frameworks that cater to different user expertise levels (5 studies). Trust was predominantly measured using quantitative surveys and questionnaires (60 studies), while the remaining studies employed qualitative methods such as interviews and user observations (30 studies).

## IV. INFERENTIAL STATISTICS

Based on the insights from Lewis & Marsh [33], many discussions around "trusting AI" shift from a simple suggestion to a directive. Instead of genuinely asking individuals to trust AI, there can be a push to make them do so. Dwyer [34] labels this as "Trust Enforcement" as opposed to "Trust Empowerment". Trust Enforcement aims to instill trust by presenting selective data, potentially portraying a certain image. In contrast, Trust Empowerment provides all relevant information, enabling individuals to determine their trust level. This distinction is important, given the personal and situational nature of trust [33]. In this research, the definition of trust encompasses two types: Trust Enforcement and Trust Empowerment.

Consequently, it is important to examine whether the existing scholarly investigations have effectively and comprehensively delineated the construct of 'trust,' while also considering the nuances of both 'Trust Enforcement' and 'Trust Empowerment'.

This study addresses the question of whether explainability in AI systems significantly enhances user trust. AI explainability is the independent variable and trust the dependent variable in this research. As per Soilemezi and Linceviciute's [35] perspective, establishing a common criterion for comparing and synthesizing independent studies can be beneficial, despite protocol confirmation, due to variations in geographic, cultural, researcher capability, data collection methods, fieldwork procedures, sample size, sampling methods, and software used [35]. In meta-analysis, effect size is a statistical concept that reflects the strength of the relationship between two variables or the magnitude of the difference between groups. It measures the impact of a factor or intervention, differing from p-values that indicate statistical significance, as effect sizes indicate real-world significance. The results are aggregated as a cumulative effect size, evaluated and validated using fixed-effects and random-effects models. Effect size is calculated separately using these models.

Table1. Study Correlations

| Model | Correlation | P-value | Z-value |
|---|---|---|---|
| Fix | 0.204 | 0.00 | 38.842 |
| Random | 0.196 | 0.00 | 17.126 |

According to Table 1 in the fixed effects model, the correlation coefficient is 0.204 with a 95% confidence interval ranging from 0.194 to 0.214, and a Z-value (measures the strength of the observed effect relative to the null hypothesis [36].) of 38.842, accompanied by a p-value (A value near zero indicates strong statistical significance in the correlation [30].) of 0.000, indicating statistical significance. These results demonstrate a positive and significant correlation between the variables under study in the fixed effects model.

In the random effects model, the correlation coefficient is 0.196 with a 95% confidence interval ranging from 0.174 to 0.218, and a Z-value of 17.126, with a p-value of 0.000, also indicating statistical significance. This model similarly shows a positive and significant correlation, although the correlation coefficient is slightly lower than in the fixed effects model. The forest plot on the right side of the figure displays the confidence intervals and correlation coefficients, indicating that both models significantly deviate from zero, signifying a notable relationship between the variables. Overall, these results indicate that explainability in AI has a positive and significant correlation with trust.

Subsequent to the primary analysis in meta-research, the heterogeneity test serves as the next step in the evaluation. This analytical tool examines the variance magnitude across the outcomes of constituent studies. It evaluates whether observed discrepancies in results stem from random variability or reflect intrinsic differences within study cohorts, procedural modalities, or resultant metrics. Pronounced heterogeneity indicates potential fallacies in amalgamating data to deduce a singular effect measure, signifying non-uniform estimations of the effect across studies.

Table2. The heterogeneity tests

| Variables | P-value | I square | Tau square | Result |
|---|---|---|---|---|
| Explainability →Trust | 0.00 | 78.337 | 0.009 | Random |

Table 2 elucidates the heterogeneity test outcomes within a meta-analytical scrutiny of the nexus between AI systems' explainability and user trust.

Thus, the meta-analysis identifies a correlation between explainability and trust in AI constructs, characterized by pronounced study heterogeneity, thereby mandating the deployment of a random-effects model to reconcile the observed disparities in effectual magnitudes. to analyze the results of the meta-analysis regarding the correlation between explainability and trust, we should consider a random-effects model.

The results of the random-effects model analysis indicate that the correlation between trust and XAI is 0.194, indicating a weak positive correlation. The 95% confidence interval for this correlation ranges from 0.174 to 0.210, suggesting that the true correlation is likely to fall within this interval with 95% certainty. The Z-value is 7.68, and the P-value is 0.00, indicating a very high level of statistical significance and confirming that the observed correlation is significantly different from zero. The forest plot visually demonstrates that the confidence intervals do not include zero, indicating a positive correlation. However, the correlation within the range of 0.174 to 0.210 is close to zero, which signifies a weak positive correlation. Correlations are typically interpreted as follows: weak correlation: values between 0.1 to 0.3; moderate correlation: values between 0.3 to 0.5; strong correlation: values between 0.5 to 1.0 [35]. Therefore, while trust increases with the increase in XAI, the increase is not substantial or strong. Thus, the analysis concludes that there is a positive and statistically significant but weak relationship between trust and XAI. Continuing with the quality assessment of a systematic review, the final and most significant issue is to examine the publication bias using the funnel plot (Figure 1).

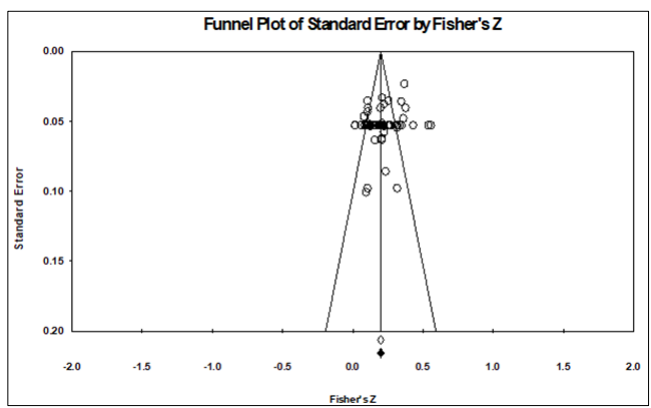

Fig. 1. Funnel Plot of Study Correlations

The symmetrical funnel plot indicates minimal publication bias, supporting the reliability of the meta-analysis results. The majority of points falling within the funnel's boundaries further suggest minimal publication bias, supporting the reliability and validity of the meta-analysis results. Integrating the empirical data from the funnel plot, the comprehensive meta-analytic table, and the forest plot, we can ascertain that a statistically significant and positive correlation exists between the explicability of artificial intelligence systems and the trust they command from users. Despite the presence of data heterogeneity and the occasional outlier within the analytical cohort, the predominant trend corroborates the proposition that an increment in the explainability of AI systems is concomitant with an elevation in user trust. Nonetheless, the precise potency of this relationship is moderate (Correlation = 0.194), suggesting that the construct of explainability, while contributory, may not be the preeminent factor influencing the trustworthiness attributed to artificial intelligence systems.

## V. DISCUSSION

Ensuring compliance with best practices and regulatory frameworks related to Trustworthy AI governance is a complex and fragmented process, involving multiple organizational units, external stakeholders, and various record-keeping systems. This fragmentation may lead to uncertainties and gaps in compliance, thereby posing reputational and regulatory risks. Moreover, specific dimensions of Trustworthy AI—such as data governance, conformance testing, quality assurance, transparency, accountability, and confidentiality—further add to this complexity. These processes often entail multiple steps, hand-offs, rework, and human oversight [37].

Meanwhile, numerous explicability techniques have been proposed without clarifying their intended purpose or are driven by questionable objectives, such as simply engendering trust. Many of these techniques rely on strong assumptions regarding the "concepts" learned by deep learning algorithms [38]. Rudin et al. [39] emphasize that "interpretable models do not necessarily engender or facilitate trust—they could also engender distrust. They simply allow users to discern whether to trust them." As a result, they advocate for the development and use of intrinsically interpretable models [39].

Recent studies indicate that explainability can indeed help build trust in AI; however, it is not the only determining factor. For example, research on the ethical and social issues surrounding AI has highlighted the significance of diversity, inclusion, bias reduction, and global inequalities in fostering user trust. These findings suggest that trust in AI cannot be achieved solely through explainability but instead requires a comprehensive and multifaceted approach [40]. Users may also feel that their identity and personal efforts are overlooked when AI makes decisions—especially positive ones—leading to decreased overall satisfaction [41].

Data localization policies can adversely affect user trust in explainable AI systems by restricting data access and reducing transparency, thereby impacting users' perceptions of security and privacy. Such policies grant governments greater oversight and control over national data, which may, in turn, negatively influence user trust in digital technologies. Beyond transparency and explainability, building trust in AI systems also requires identifying and managing biases, harmful behaviors, and other risks associated with the use of large language models (LLMs). Researchers must carefully assess potential risks, limitations, and biases embedded in AI models and adopt value-sensitive design methods to develop trustworthy and explainable AI systems [42]. By doing so, AI models not only maintain accuracy and transparency but also address potential biases and ethical concerns. Integrating these practices into AI development can enhance the overall reliability and credibility of such systems [43].

Algorithmic bias has serious implications for inclusion and diversity, posing significant ethical challenges. There is an urgent need for globally oriented research to examine ethical issues in AI design and development. Studies suggest that data colonialism, particularly in the Global South, exacerbates digital inequalities and further marginalizes vulnerable populations. Current ethical design practices appear insufficient; hence, researchers are encouraged to employ practical approaches to ethics that consider not only the design and development phases but also the broader socio-technical systems, treating ethics as an ongoing practical activity.

Building trust in AI remains an evolving endeavor, demanding continuous efforts to establish and sustain trust at every phase of AI development and deployment [44]. Although there is no universally accepted definition of trustworthy AI, attention has converged on eight foundational principles widely recognized in the field: accountability, fairness and non-discrimination, human oversight of technology, privacy, professional responsibility, the enhancement of human values, safety and security, and transparency and explainability [45]. To strengthen trust in AI, organizations and developers are encouraged to measure their adherence to these principles throughout system creation [46].

One key challenge within the domain of XAI is the lack of a coherent conceptualization of "understanding." In philosophy, three differing views emerge: the first insists that genuine understanding can only be achieved through accurate explicability [47], [48]; the second suggests alternate methods that may involve inaccurate explicability [48]; and the third, more moderate stance, allows for the inclusion of certain inaccuracies within explicability [49]. Given these perspectives, there is no consensus on which approach best applies to XAI, resulting in diverse interpretations of how AI "understanding" should be articulated. Contributions from philosophical and psychological discussions could significantly advance how we conceive and implement XAI [20].

In investigating the factors that influence users' trust in AI, it is critical to consider all three philosophical perspectives. Each offers valuable insights into the complexities of ethical decision-making and provides a more holistic understanding of how trust in AI may be cultivated. Unlike many previous studies, which either adopted a purely qualitative approach or were confined to specific case studies, this research employs meta-analysis to draw upon multiple quantitative datasets, thereby offering a comprehensive evaluation of the actual strength of the relationship between explainability and trust.

Our findings reveal that while explainability positively influences trust, the correlation coefficient of 0.194 is notably weaker than what some earlier studies had suggested. Consequently, this study underscores the importance of addressing additional factors—such as ethical and social considerations—while demonstrating that an overemphasis on explainability alone may obscure deeper, more fundamental sources of user distrust in AI systems. This insight can guide policymakers and system designers to implement more effective measures for risk management, bias mitigation, and the accommodation of cultural concerns, going beyond technical transparency of models. By offering a meta-analytical perspective on the literature, this research paves the way for future studies to systematically identify and quantify those complementary factors that can enhance or diminish the role of explainability in building user trust.

Trust can be conceptualized as the decision to rely on something that cannot be fully understood, predicted, or controlled, especially when risk is involved [33]. In contrast, trust refers to the belief or confidence that a stakeholder places in a system. Trustworthiness, on the other hand, is an objective characteristic intrinsic to the system itself, while trust remains inherently subjective. Various interpretations and definitions of trustworthiness abound in the literature, as identified in numerous references [50]. Trustworthiness closely aligns with the notion of relying on a dependable system. Practically, it is more fruitful for designers to focus on elements they can directly control rather than external factors. Consequently, concentrating on enhancing the trustworthiness of a system may be more achievable than seeking to manipulate subjective trust. Trustworthiness is linked to the system's inherent qualities, whereas trust reflects a stakeholder's subjective choice [51], [52]. If a system's trustworthiness is neglected and proves unreliable, two undesired outcomes may ensue: stakeholders might lose trust in an undependable system, or conversely, they could mistakenly trust an inconsistent one, resulting in potentially harmful consequences. In either scenario, introducing a trustworthy system is both ethically advisable and strategically sound, even if immediate user trust does not always follow.

In addition to these ethical and design implications, the findings of this study highlight the pivotal role of explainability in tackling broader socio-technical issues. For instance, incorporating explainable AI into healthcare, justice, or governance systems can promote transparency and reduce biases that disproportionately impact marginalized groups, thereby fostering greater trust. However, this research also shows that explainability alone is insufficient; it should be accompanied by a nuanced understanding of the socio-cultural contexts in which AI operates to ensure inclusivity, fairness, and ethical accountability. Viewing AI explainability within a socio-technical framework thus offers a pathway for developing systems aligned with both human values and societal needs, reinforcing trust as a collective, enduring goal.

## VI. Conclusion

The objective of this investigation was to examine the relationship between the explainability of artificial intelligence and trust. We employed a meta-analytical approach and conducted a comprehensive review of pertinent studies. After analyzing 90 academic articles that met the predefined criteria established by the researchers, verified using meta-analysis software, and reported relevant statistical indicators, we found a statistically significant but low positive correlation between the explainability of artificial intelligence and trust. Specifically, the correlation coefficient of 0.194 underscores that while explainability does indeed bolster trust, its impact remains modest in practice. Consequently, focusing solely on explainability may not suffice, and complementary factors—such as ethical safeguards, user engagement, and domain-specific considerations—should be integrated to achieve a more robust and sustainable level of trust in AI systems. Therefore, it is evident that while explainability plays a role in fostering trust, it alone is insufficient. A multifaceted approach that includes ethical considerations, bias management, and socio-technical understanding can contribute to building robust and trustworthy AI systems. Consequently, this research illustrates that developers and policymakers must, while prioritizing model explainability, pay greater attention to ethical, cultural, and social requirements. Combining explainability approaches with value-sensitive design and proactive risk management can enhance user confidence and acceptance of AI systems, ultimately leading to greater efficiency, fairness, and accountability in intelligent systems.

### Acknowledgement

This research was undertaken, in part, thanks to funding from cite the Canada Research Chairs Program.

## BIOGRAPHICAL NOTES

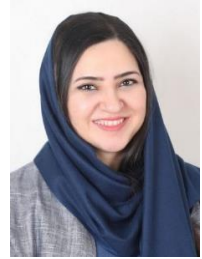

**Dr. Zahra Atf** currently serving as a Visiting Researcher at the Trustworthy Artificial Intelligence Lab at Ontario Tech University. She holds a Ph.D. in Business with a specialization in Marketing and a Master's degree in Information Systems Management and Industrial engineering.

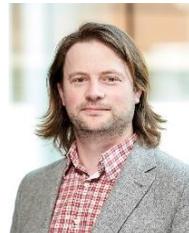

**Dr. Peter R. Lewis** holds a Canada Research Chair in Trustworthy Artificial Intelligence (AI), at Ontario Tech University, Canada, where he is an Associate Professor and Director of the Trustworthy AI Lab. Peter's research advances both foundational and applied aspects of AI and draws on extensive experience applying AI commercially and in the non-profit sector. He is interested in where AI meets society, and how to help that relationship work well. His current research is concerned with challenges of trust, bias, and accessibility in AI, as well as how to create more socially intelligent AI systems, such that they work well as part of society, explicitly taking into account human factors such as norms, values, social action, and trust. He has a PhD in Computer Science from the University of Birmingham, UK.